\documentclass[10pt,twocolumn,letterpaper]{article}

\usepackage{iccv}

\usepackage{times}
\usepackage{epsfig}
\usepackage{graphicx}
\usepackage{grffile}
\usepackage{amsmath}
\usepackage{amssymb}
\usepackage{booktabs}
\usepackage{mwe}
\usepackage{colortbl}
\usepackage{multirow}
\usepackage[ruled,vlined]{algorithm2e}

\usepackage{enumitem}
\setlist{nosep}

\usepackage{color}

\definecolor{DarkGreen}{rgb}{0.0, 0.6, 0.2}
\definecolor{DarkerGreen}{rgb}{0.0, 0.4, 0.1}

\definecolor{DarkRed}{rgb}{0.5, 0.1, 0.1}
\definecolor{DarkerRed}{rgb}{0.5, 0.1, 0.0}

\usepackage[breaklinks=true,bookmarks=false,colorlinks=true,linkcolor=red,citecolor=blue]{hyperref}

\iccvfinalcopy % *** Uncomment this line for the final submission

\def\iccvPaperID{****} % *** Enter the ICCV Paper ID here

\ificcvfinal\fi

\begin{document}

%%%%%%%%% TITLE

\title{1st Place Solution for the UVO Challenge on Image-based Open-World Segmentation 2021\\
%\Large \textmd{Team: Alpes\_Runner}
}

\author{Yuming Du$^1$ \qquad \quad Wen Guo$^2$ \qquad \quad Yang Xiao$^1$ \qquad \quad Vincent Lepetit$^1$\\
$^1$~LIGM, Ecole des Ponts, Univ Gustave Eiffel, CNRS, Marne-la-Vallée, France\\
$^2$~Inria, Univ. Grenoble Alpes, CNRS, Grenoble INP, LJK, 38000 Grenoble, France\\
{\tt\small \{yuming.du, yang.xiao, vincent.lepetit\}@enpc.fr}, {\tt\small wen.guo@inria.fr}\\
\url{https://github.com/dulucas/UVO_Challenge}
}

\maketitle

%%%%%%%%% ABSTRACT
\begin{abstract}
We describe our two-stage instance segmentation framework we use to compete in the challenge. The first stage of our framework consists of an object detector, which generates object proposals in the format of bounding boxes. Then, the images and the detected bounding boxes are fed to the second stage, where a segmentation network is applied to segment the objects in the bounding boxes. We train all our networks in a class-agnostic way. Our approach achieves the first place in the UVO 2021 Image-based Open-World Segmentation Challenge.
\end{abstract}

%%%%%%%%% BODY TEXT
\section{Method}
\label{sec:method}
In this section, we present our method for open-world instance segmentation. Our method consists of two stages, the first stage is an anchor-based object detection network that predicts bounding boxes for objects in the images. The second stage consists of a segmentation network, which takes the image and the bounding box as input and segments the object in the bounding box. Thanks to this architecture, we can train the detection network and segmentation network separately on different datasets, instead of following an ``end-to-end'' training approach. The large input size of segmentation network results in finer and better mask predictions.

% Compared with end-to-end instance segmentation network, the advantage of our method is that we can train the detection network and segmentation network separately on different datasets. And the large input size of segmentation network results in finer and better mask predictions.

\subsection{Detection}
\begin{figure*}
\begin{center}
  \begin{tabular}{cccc}
    \includegraphics[width=1.0\linewidth]{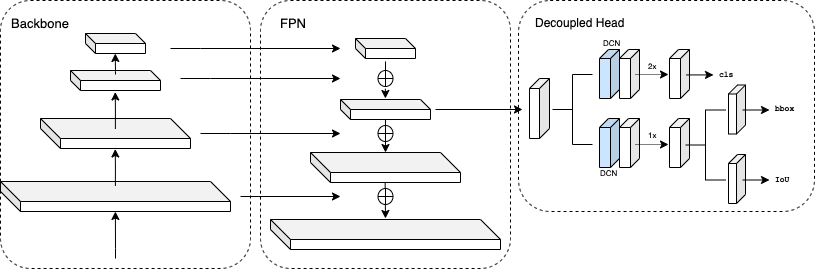}
\end{tabular}
\end{center}
\caption{\label{fig:detection network} Overview of our detection network architecture. DCN stands for Deformable Convolutional neural Network\cite{dai2017deformable}.}
\end{figure*}

In this section, we introduce the object detection network we used and the techniques adopted that help to improve the recall of the object detector. An overview of our network architecture is shown in Figure~\ref{fig:detection network}.

\paragraph{Baseline.} We adopt the Region Proposal Network~(RPN) of \cite{ren2015faster} as our baseline network. Our backbone network consists of a ResNet-50\cite{he2016deep} network with the Feature Pyramid Network~(FPN) of  \cite{lin2017feature} for multi-scale feature extraction. A classification head and a regression head are used for predicting the ``objectness'' of the region contents and for bounding box regression, respectively.

\paragraph{Cascade Region Proposal Network~(RPN).} To further improve the quality of the predicted object proposals, as in \cite{vu2019cascade}, we adopt a two-stage cascade architecture for proposal generation. Only one single anchor is used for each location in the feature maps. The predicted bounding boxes from the first stage are used as the input for the second stage. Adaptive convolution is used to solve the misalignment between the anchors and the features throughout the stage. The Focal loss~\cite{lin2017focal} is used for classification and GIoU loss~\cite{rezatofighi2019generalized} for bounding box regression.

% \vincentrmk{Does it make sense to use the focal loss if you have only 2 classes, object vs non-object?} \dymrmk{Yes, the focal loss aims to balance the positive/negative samples, as during the training of a detector, usually there are just a few positive samples(around the ground truth bounding boxes) with a lot of negative samples. One strategy is to random sample a fixed number of pos/neg samples with a fixed ratio(as done in Faster R-CNN etc), as in our case, all pos/neg samples are used for training, focal loss can help a lot(this is not novel, what we did is similar to ATSS\cite{zhang2020bridging}).}

\paragraph{IoU Branch.} In addition to the classification and bounding box branches, we use an IoU branch, which predicts the IoU between the predicted bounding boxes and the ground truth bounding boxes. During inference, the objectness score is calculated as the geometric mean of the predicted IoU score and the classification score, as also done in \cite{jiang2018acquisition, ge2021yolox} for example.

% \vincentrmk{Is this novel?  Can you justify this better?} \dymrmk{This is not novel. The IoU branch has been widely used in previous works\cite{ge2021yolox, jiang2018acquisition}.} \dym{For each bounding box proposal, our detection network will predict two scores, one for foreground/background classfication and the other for IoU score which represents the IoU between the predicted bounding box with ground truth bounding box. We calculate the geometric mean of these two scores and use it as the final objectness scores for the predicted bounding boxes.}

\paragraph{Decoupled Heads. } Decoupled heads have been widely used in previous works~\cite{ge2021yolox, tian2019fcos} and have been demonstrated to be effective to ease the conflict between the classification and regression tasks in object detection. We adopt decoupled heads in our network. Heads across all pyramid levels share the same weights to save memory. We further replace the first convolutional layer of the decoupled heads with deformable convolutional layers~\cite{dai2017deformable} and we will show in the later section that this could further improve our results.

\paragraph{Proposal Sampling. } Label assignment aims to define positive/negative samples during training. We replace the IoU-based label assignment with the SimOTA label assignment~\cite{ge2021yolox} in our network. SimOTA is a simplified version of OTA label assignment~\cite{ge2021ota}, which formulates the label assignment as an Optimal Transport~(OT) problem and finds the negative/positive samples by measuring their transportation costs to ground truth bounding boxes. Instead of using the Sinkhorn-Knopp algorithm, SimOTA simply takes the top-K candidates that are centered at the center of the ground truth bounding boxes. In addition to this, we adopt different negative/positive label assignment strategies for classification head and regression head.
%\vincentrmk{I dont get it, what are these strategies?} \dymrmk{The strategy refers to SimOTA that we use for pos/neg sampling. Instead of using one SimOTA for both classification and regression, we use 2 SimOTAs with different hyperparameters for classification and regression respectively. Compared to the SimOTA for classification, the one for regression assign more positive samples for candidate anchors during training.}. 
In particular, instead of using one SimOTA for both classification and regression, we use two SimOTAs with different hyperparameters for classification and regression respectively. Compared to the SimOTA for classification, the one for regression assign more positive samples for candidate anchors during training.
For the regression head, we loose the selection criterion to generate more positive samples.

\paragraph{Non-maximum suppression~(NMS). } Non-maximum suppression is used to remove duplicate bounding boxes. We set our NMS threshold to 0.8.

\paragraph{Feature Pyramid Network~(FPN). } We further add CARAFE~\cite{wang2019carafe} blocks in our FPN~\cite{lin2017feature} network for a better feature upsampling.

\paragraph{Backbone. } We use the recent Swin-L transformer~\cite{liu2021swin} as our backbone network.

%\paragraph{SWA. } We further rely on the SWA strategy~\cite{zhang2020swa} for model ensemble: We average the weights of our model over the last 3 epochs of our model to obtain the final object detection model.

% We take the last 3 epochs of our model and then average their weights to get the final object detection model.

\subsection{Segmentation}
The segmentation network we used for this challenge is based on the Upernet~\cite{xiao2018unified} architecture and the Swin-L transformer~\cite{liu2021swin} as the backbone network. The input consists of images and predicted bounding boxes, the bounding boxes are first used to crop the images to image patches. Then, the image patches are all resized to 512$\times$512 regardless of their height/width ratios.
%-------------------------------------------------------------------------

\section{Dataset}
\label{sec:dataset}
\subsection{Detection}
ImageNet 22k is used to pre-train our backbone network. We then train our detectors on the COCO dataset~\cite{lin2014microsoft}. Finally, the pre-trained detectors are fine-tuned on the UVO-Sparse dataset and the UVO-Dense dataset~\cite{wang2021unidentified}.

\subsection{Segmentation} ImageNet 22k~\cite{imagenet_cvpr09} is used to pre-train our backbone network. We then train our segmentation network on a combination of the OpenImage~\cite{OpenImages2}, PASCALVOC~\cite{everingham2010pascal}, and COCO~\cite{lin2014microsoft} datasets. Finally, the pre-trained segmentation networks are fine-tuned on the UVO-Sparse dataset and the UVO-Dense dataset~\cite{wang2021unidentified}.
%-------------------------------------------------------------------------

\begin{figure*}
\begin{center}
  \begin{tabular}{cccc}
    \includegraphics[width=1.0\linewidth]{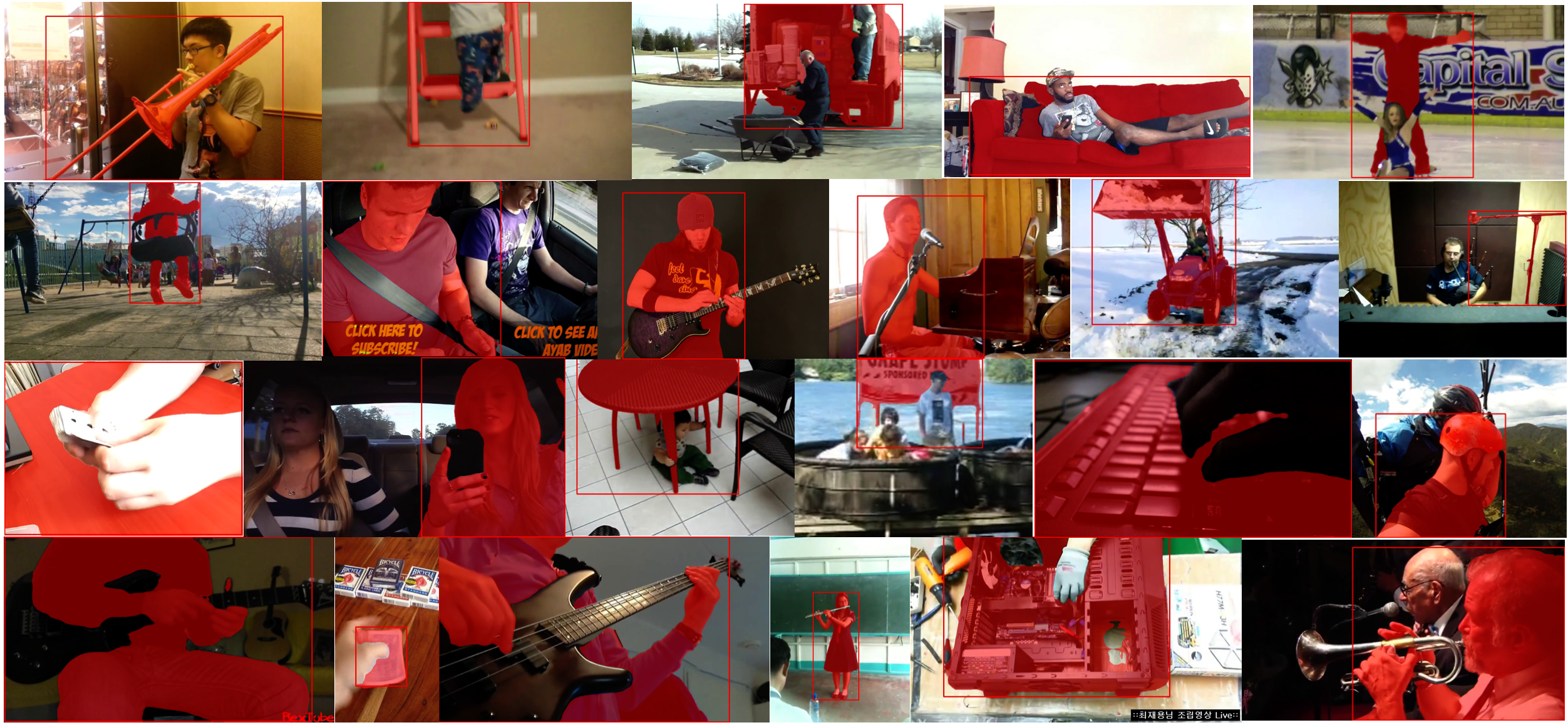}
\end{tabular}
\end{center}
\caption{\label{fig:seg example} Some of our predictions on the UVO-Sparse test dataset. The predicted bounding boxes and segmented objects are shown in red. Our segmentation network can segment the object regardless the motion blur, strong occlusion, extreme pose etc.}
\end{figure*}

\section{Implementation Details}
\label{sec:implementation details}
\subsection{Detection}
We use MMDetection~\cite{mmdetection} to train our detectors. For the backbone network, we get the Swin-L transformer pre-trained on ImageNet 22k from \footnote{\url{https://github.com/microsoft/Swin-Transformer}}. All our detectors are trained with Detectron ‘1x’ setting. For data augmentation, we use the basic data augmentation strategy as in \cite{he2017mask} for all experiments. The center ratio of both SimOTA samplers are set to 0.25, the top-K number for classification head is set to 10, while the top-K number for regression head is set to 20 to involve more positive samples. Four 3$\times$3 conv layers are used in the classification branch and the regression head, IoU branch shares the same conv layers with the regression branch. To train the detector with Swin-L transformer backbone, we adopt the AdamW as the optimizer and set the init learning rate to 1e-4. The batch size is set to 16. After training on COCO, we fine-tune the detector on the combination of UVO-Sparse dataset and UVO-Dense dataset for 6 epochs. All our detectors are trained in the class-agnostic way. Test time augmentation is used during inference to further boost the network performance.

\subsection{Segmentation}
We use MMSegmentation~\cite{mmseg2020} to train our segmentation network. We use the same backbone network as our detection network. During training, given an image and an instance mask, we first generate a bounding box that envelopes the instance mask, then a 20 pixel margin is added to the bounding box in all directions. We use the generated bounding box to crop the image and resize the image patch to 512$\times$512. Random flipping, random photometric distortion, and random bounding box jitter are used as data augmentation. We adopt 'poly' learning rate policy and set the initial learning rate to 6e-5. The batch size is set to 32 and AdamW\cite{loshchilov2017decoupled} is used as the optimizer. We first train our network on the combination of the OpenImage~\cite{OpenImages2}, PASCALVOC~\cite{everingham2010pascal} and COCO~\cite{lin2014microsoft} datasets for 300k iterations, then we finetune the network on the combination of the UVO-Dense and UVO-Sparse datasets for 100k iterations with initial learning rate set to 6e-6. All our segmentation networks are trained in a class-agnostic way, thus, segmenting the object in the cropped path becomes a foreground/background segmentation problem. Only flip test augmentation is adopted during inference.

%-------------------------------------------------------------------------
\begin{table}[!t]
  \addtolength{\tabcolsep}{-1.pt}
  \begin{center}
  \scalebox{.9}
	     {
	       \begin{tabular}{@{}r | ccc @{}}
	         \toprule
	          & $AR@100$ & $AR@300$ & $AR@1000$ \\ 
	         \midrule
	         Baseline~(Res50-FPN RPN)  
	         & 44.6 & 52.9 & 58.3 \\
	         
	         \small {+ Cascasde RPN}
	         & 61.1 & 67.6 & 71.7 \\
	         
	         \small {+ SimOTA + IoU Branch}
	         & 62.6 & 68.1 & 71.7 \\
	         
	         \small {+ Decoupled Head}
	         & 64.5 & 69.8 & 73.1 \\
	         
	         \small {+ Two samplers}
	         & 64.8 & 70.1 & 73.6 \\
	         
	         \small {+ CARAFE}
	         & 65.2 & 70.4 & 73.9 \\
	         
	         \small {+ DCNv2}
	         & 65.6 & 70.8 & 74.2 \\
	         
	         %\scriptsize {+ Swin-S}
	         %& 68.7 & 73.1 & 76.0 \\
	         
	         \small {+ Swin-L}
	         & 70.7 & 74.9 & 77.4 \\
	         
	         \bottomrule
	       \end{tabular}
	     }
  \end{center}
%   \vspace{-3mm}
  \caption{{\bf Ablation study} on the different components of our method on COCO \textit{val2017}.
  }
  \label{tab:overall ablation}
\end{table}

\begin{table*}[!t]
  \addtolength{\tabcolsep}{-1.pt}
  \begin{center}
  \scalebox{1.1}
	     {
	       \begin{tabular}{@{}l | cccccc @{}}
	         \toprule
	          Teams & $AR@100$ & $AP$ & $AP@.5$ & $AP@.75$ & $AR@1$ & $AR@10$ \\ 
	         \midrule
	         lll\_uvo
	         & 33.39 & 22.60 & 41.23 & 22.08 & 7.49 & 27.49 \\
	         
	         FUFU
	         & 36.58 & 21.43 & 41.04 & 20.62 & 6.83 & 26.63 \\
	         
	         FUFUUVO
	         & 38.36 & 21.47 & 42.24 & 20.44 & 6.42 & 26.37 \\

	         CASIT\_UVO
	         & 39.33 & 21.59 & 42.32 & 20.38 & 6.45 & 26.37 \\

	         Jinzhao\_Zhou(80\_1)
	         & 39.91 & 22.30 & 43.76 & 20.77 & 6.86 & 27.45 \\
	         
	         Baseline by host
	         & 41.43 & 23.62 & 45.06 & 22.83 & 6.98 & 27.76 \\
	         
	         WRTC
	         & 58.06 & 39.22 & 64.07 & 41.62 & 9.17 & 40.05 \\
	         
	         UAT
	         & 60.61 & 38.40 & 62.17 & 40.86 & 9.40 & 41.32 \\
	         
	         Ours
	         & \textbf{61.77} & \textbf{39.73} & \textbf{59.87} & \textbf{42.95} & \textbf{9.71} & \textbf{40.61} \\
	         
	         \bottomrule
	       \end{tabular}
	     }
  \end{center}
%   \vspace{-3mm}
  \caption{{\bf Challenge final results} on the UVO-Sparse test dataset. Our method surpasses the baseline and other teams by a large margin. }
  \label{tab:overall ablation}
\end{table*}

\section{Ablation Study}
\label{sec:ablation study}
\subsection{Detection}
In this section, we ablate the different components in our detection network. We train and evaluate our detectors using COCO \textit{train2017/val2017}. The results are shown in Table~\ref{tab:overall ablation}.

\section{Visualization}
We visualize the results predicted by our segmentation network given images and bounding boxes in Figure~\ref{fig:seg example}. Our method results in high quality mask predictions.

%-------------------------------------------------------------------------
\section{Potential Improvements}
\paragraph{More training data.} We only pre-train our detectors on the COCO dataset, while recent works\cite{shao2019objects365} show that pre-training on some large datasets like \cite{shao2019objects365, OpenImages2} can further improve the performance of detectors. With more classes and more objects of different shapes and geometries, the detectors might learn a better representation which helps to find more objects and better localize them.

\paragraph{Data augmentation.} Our detectors are all trained using naive data augmentation like resize, flip, etc. Recent works~\cite{ghiasi2021simple, fang2019instaboost, cubuk2018autoaugment} show that object detectors can largely benefit from strong data augmentation. Adding data augmentation might further improve the performance of our network.

\paragraph{Misc.} Some hyperparameters are still chosen arbitrarily, including the number of training epochs for fine-tuning on the UVO-Sparse and UVO-Dense datasets and the hyperparameters for label assignment. Tuning these hyperparameters could lead to better performance.

{\small
\bibliographystyle{ieee_fullname}
\bibliography{egbib}

\begin{thebibliography}{10}\itemsep=-1pt

\bibitem{mmdetection}
Kai Chen, Jiaqi Wang, Jiangmiao Pang, Yuhang Cao, Yu Xiong, Xiaoxiao Li,
  Shuyang Sun, Wansen Feng, Ziwei Liu, Jiarui Xu, Zheng Zhang, Dazhi Cheng,
  Chenchen Zhu, Tianheng Cheng, Qijie Zhao, Buyu Li, Xin Lu, Rui Zhu, Yue Wu,
  Jifeng Dai, Jingdong Wang, Jianping Shi, Wanli Ouyang, Chen~Change Loy, and
  Dahua Lin.
\newblock {MMDetection}: Open mmlab detection toolbox and benchmark.
\newblock {\em arXiv preprint arXiv:1906.07155}, 2019.

\bibitem{mmseg2020}
MMSegmentation Contributors.
\newblock {MMSegmentation}: Openmmlab semantic segmentation toolbox and
  benchmark.
\newblock \url{https://github.com/open-mmlab/mmsegmentation}, 2020.

\bibitem{cubuk2018autoaugment}
Ekin~D Cubuk, Barret Zoph, Dandelion Mane, Vijay Vasudevan, and Quoc~V Le.
\newblock Autoaugment: Learning augmentation policies from data.
\newblock {\em arXiv preprint arXiv:1805.09501}, 2018.

\bibitem{dai2017deformable}
Jifeng Dai, Haozhi Qi, Yuwen Xiong, Yi Li, Guodong Zhang, Han Hu, and Yichen
  Wei.
\newblock Deformable convolutional networks.
\newblock In {\em Proceedings of the IEEE international conference on computer
  vision}, pages 764--773, 2017.

\bibitem{imagenet_cvpr09}
J. Deng, W. Dong, R. Socher, L.-J. Li, K. Li, and L. Fei-Fei.
\newblock {ImageNet: A Large-Scale Hierarchical Image Database}.
\newblock In {\em CVPR09}, 2009.

\bibitem{everingham2010pascal}
Mark Everingham, Luc Van~Gool, Christopher~KI Williams, John Winn, and Andrew
  Zisserman.
\newblock The pascal visual object classes (voc) challenge.
\newblock {\em International journal of computer vision}, 88(2):303--338, 2010.

\bibitem{fang2019instaboost}
Hao-Shu Fang, Jianhua Sun, Runzhong Wang, Minghao Gou, Yong-Lu Li, and Cewu Lu.
\newblock Instaboost: Boosting instance segmentation via probability map guided
  copy-pasting.
\newblock In {\em Proceedings of the IEEE/CVF International Conference on
  Computer Vision}, pages 682--691, 2019.

\bibitem{ge2021ota}
Zheng Ge, Songtao Liu, Zeming Li, Osamu Yoshie, and Jian Sun.
\newblock Ota: Optimal transport assignment for object detection.
\newblock In {\em Proceedings of the IEEE/CVF Conference on Computer Vision and
  Pattern Recognition}, pages 303--312, 2021.

\bibitem{ge2021yolox}
Zheng Ge, Songtao Liu, Feng Wang, Zeming Li, and Jian Sun.
\newblock Yolox: Exceeding yolo series in 2021.
\newblock {\em arXiv preprint arXiv:2107.08430}, 2021.

\bibitem{ghiasi2021simple}
Golnaz Ghiasi, Yin Cui, Aravind Srinivas, Rui Qian, Tsung-Yi Lin, Ekin~D Cubuk,
  Quoc~V Le, and Barret Zoph.
\newblock Simple copy-paste is a strong data augmentation method for instance
  segmentation.
\newblock In {\em Proceedings of the IEEE/CVF Conference on Computer Vision and
  Pattern Recognition}, pages 2918--2928, 2021.

\bibitem{he2017mask}
Kaiming He, Georgia Gkioxari, Piotr Doll{\'a}r, and Ross Girshick.
\newblock Mask r-cnn.
\newblock In {\em Proceedings of the IEEE international conference on computer
  vision}, pages 2961--2969, 2017.

\bibitem{he2016deep}
Kaiming He, Xiangyu Zhang, Shaoqing Ren, and Jian Sun.
\newblock Deep residual learning for image recognition.
\newblock In {\em Proceedings of the IEEE conference on computer vision and
  pattern recognition}, pages 770--778, 2016.

\bibitem{jiang2018acquisition}
Borui Jiang, Ruixuan Luo, Jiayuan Mao, Tete Xiao, and Yuning Jiang.
\newblock Acquisition of localization confidence for accurate object detection.
\newblock In {\em Proceedings of the European conference on computer vision
  (ECCV)}, pages 784--799, 2018.

\bibitem{OpenImages2}
Ivan Krasin, Tom Duerig, Neil Alldrin, Vittorio Ferrari, Sami Abu-El-Haija,
  Alina Kuznetsova, Hassan Rom, Jasper Uijlings, Stefan Popov, Shahab Kamali,
  Matteo Malloci, Jordi Pont-Tuset, Andreas Veit, Serge Belongie, Victor Gomes,
  Abhinav Gupta, Chen Sun, Gal Chechik, David Cai, Zheyun Feng, Dhyanesh
  Narayanan, and Kevin Murphy.
\newblock Openimages: A public dataset for large-scale multi-label and
  multi-class image classification.
\newblock {\em Dataset available from
  https://storage.googleapis.com/openimages/web/index.html}, 2017.

\bibitem{lin2017feature}
Tsung-Yi Lin, Piotr Doll{\'a}r, Ross Girshick, Kaiming He, Bharath Hariharan,
  and Serge Belongie.
\newblock Feature pyramid networks for object detection.
\newblock In {\em Proceedings of the IEEE conference on computer vision and
  pattern recognition}, pages 2117--2125, 2017.

\bibitem{lin2017focal}
Tsung-Yi Lin, Priya Goyal, Ross Girshick, Kaiming He, and Piotr Doll{\'a}r.
\newblock Focal loss for dense object detection.
\newblock In {\em Proceedings of the IEEE international conference on computer
  vision}, pages 2980--2988, 2017.

\bibitem{lin2014microsoft}
Tsung-Yi Lin, Michael Maire, Serge Belongie, James Hays, Pietro Perona, Deva
  Ramanan, Piotr Doll{\'a}r, and C~Lawrence Zitnick.
\newblock Microsoft coco: Common objects in context.
\newblock In {\em European conference on computer vision}, pages 740--755.
  Springer, 2014.

\bibitem{liu2021swin}
Ze Liu, Yutong Lin, Yue Cao, Han Hu, Yixuan Wei, Zheng Zhang, Stephen Lin, and
  Baining Guo.
\newblock Swin transformer: Hierarchical vision transformer using shifted
  windows.
\newblock {\em arXiv preprint arXiv:2103.14030}, 2021.

\bibitem{loshchilov2017decoupled}
Ilya Loshchilov and Frank Hutter.
\newblock Decoupled weight decay regularization.
\newblock {\em arXiv preprint arXiv:1711.05101}, 2017.

\bibitem{ren2015faster}
Shaoqing Ren, Kaiming He, Ross Girshick, and Jian Sun.
\newblock Faster r-cnn: Towards real-time object detection with region proposal
  networks.
\newblock {\em Advances in neural information processing systems}, 28:91--99,
  2015.

\bibitem{rezatofighi2019generalized}
Hamid Rezatofighi, Nathan Tsoi, JunYoung Gwak, Amir Sadeghian, Ian Reid, and
  Silvio Savarese.
\newblock Generalized intersection over union: A metric and a loss for bounding
  box regression.
\newblock In {\em Proceedings of the IEEE/CVF Conference on Computer Vision and
  Pattern Recognition}, pages 658--666, 2019.

\bibitem{shao2019objects365}
Shuai Shao, Zeming Li, Tianyuan Zhang, Chao Peng, Gang Yu, Xiangyu Zhang, Jing
  Li, and Jian Sun.
\newblock Objects365: A large-scale, high-quality dataset for object detection.
\newblock In {\em Proceedings of the IEEE/CVF International Conference on
  Computer Vision}, pages 8430--8439, 2019.

\bibitem{tian2019fcos}
Zhi Tian, Chunhua Shen, Hao Chen, and Tong He.
\newblock Fcos: Fully convolutional one-stage object detection.
\newblock In {\em Proceedings of the IEEE/CVF international conference on
  computer vision}, pages 9627--9636, 2019.

\bibitem{vu2019cascade}
Thang Vu, Hyunjun Jang, Trung~X Pham, and Chang~D Yoo.
\newblock Cascade rpn: Delving into high-quality region proposal network with
  adaptive convolution.
\newblock {\em arXiv preprint arXiv:1909.06720}, 2019.

\bibitem{wang2019carafe}
Jiaqi Wang, Kai Chen, Rui Xu, Ziwei Liu, Chen~Change Loy, and Dahua Lin.
\newblock Carafe: Content-aware reassembly of features.
\newblock In {\em Proceedings of the IEEE/CVF International Conference on
  Computer Vision}, pages 3007--3016, 2019.

\bibitem{wang2021unidentified}
Weiyao Wang, Matt Feiszli, Heng Wang, and Du Tran.
\newblock Unidentified video objects: A benchmark for dense, open-world
  segmentation.
\newblock {\em arXiv preprint arXiv:2104.04691}, 2021.

\bibitem{xiao2018unified}
Tete Xiao, Yingcheng Liu, Bolei Zhou, Yuning Jiang, and Jian Sun.
\newblock Unified perceptual parsing for scene understanding.
\newblock In {\em Proceedings of the European Conference on Computer Vision
  (ECCV)}, pages 418--434, 2018.

\end{thebibliography}
}

\end{document}